\documentclass[runningheads]{llncs}
\usepackage{graphicx}
\usepackage{amssymb}
\usepackage[cmex10]{amsmath}
\usepackage{color}
\usepackage{comment}
\begin{document}
\title{Emotion Recognition with Spatial Attention and Temporal Softmax Pooling}
%
%
\author{Masih Aminbeidokhti \and
Marco Pedersoli \and
Patrick Cardinal \and Eric Granger}
\authorrunning{M. Aminbeidokhti et al.}
%
\institute{Laboratoire d'imagerie, de vision et d'intelligence artificielle (LIVIA) \\ 
\'Ecole de technologie sup\'erieure, Montreal, Canada \\  
\email{masih.aminbeidokhti.1@ens.etsmtl.ca, \{marco.pedersoli, patrick.cardinal, eric.granger\}@etsmtl.ca}}
\maketitle              
\begin{abstract}
Video-based emotion recognition is a challenging task because it requires to distinguish the small deformations of the human face that represent emotions, while being invariant to stronger visual differences due to different identities. State-of-the-art methods normally use complex deep learning models such as recurrent neural networks (RNNs, LSTMs, GRUs), convolutional neural networks (CNNs, C3D, residual networks) and their combination. 
In this paper, we propose a simpler approach that combines a CNN pre-trained on a public dataset of facial images with (1) a spatial attention mechanism, to localize the most important regions of the face for a given emotion, and (2) temporal softmax pooling, to select the most important frames of the given video. Results on the challenging EmotiW dataset show that this approach can achieve higher accuracy than more complex approaches.

\keywords{Affective Computing \and Emotion Recognition \and Attention Mechanisms \and Convolutional Neural Networks.}
\end{abstract}

\section{Introduction}

Designing a system capable of encoding discriminant features for video-based emotion recognition is challenging because the appearance of faces may vary considerably according to the specific subject, capture conditions (pose, illumination, blur), and sensors. It is difficult to encode common and discriminant spatio-temporal features of emotions while suppressing these context- and subject-specific facial variations.  

Recently, emotion recognition has attracted attention from the computer vision community because state-of-the-art methods are finally providing results that are comparable with human performance. Thus, these methods are now becoming more reliable, are beginning to be deployed in real-world applications \cite{911197}. However, at this point, it is not yet clear what is the right recipe of success in terms of machine learning architectures. Several state-of-the-art methods \cite{knyazev2018leveraging},\cite{Liu:2018:MBE:3242969.3264989} originating from challenges in which multiple teams provide results on the same benchmark without having access training-set annotations. Although these challenges measure improvements in the field. one a drawback of challenges is that result focuses mostly on final accuracy of approaches, without taking into account other factors such as their computational cost, architectural complexity, quantity of hyper-parameters to tune, versatility, generality of the approach, etc. As a consequence, there is no clear cost-benefit analysis for component appearing in top-performing methods and often represent complex deep learning architectures.

In this paper, we aim to shed some light on these issues by proposing a simple approach for emotion recognition that i) is based on the very well-known VGG16 network which is pre-trained on face images; ii) has a very simple yet performing mechanism to aggregate temporal information; and iii) uses an attention model to select which part of the face is the most important to recognize a certain emotion. 
For the selection of the approach to use, we show that a basic convolutional neural network such as VGG can perform as well or even better than more complex models when pre-trained on clean data.
For temporal aggregation, we show that softmax pooling is an excellent way to select information from different frames because it is a generalization of max and average pooling. Additionally, in contrast to more complex techniques (e.g. attention), it does not require additional sub-networks and therefore additional parameters to train, which can easily lead to overfitting when dealing with relatively small datasets, a common problem in this field. Finally, we show that for the selection of the most discriminative parts of a face for recognizing an emotion, an attention mechanism is necessary to improve performance. For doing that, we built a small network with multiple attention heads \cite{lin2017structured} that can simultaneously focus on different parts of a human face.

The rest of the paper is organized as follows. The next section described related work. Then, our methods based on spatial attention and temporal softmax are presented. Finally, in our experimental evaluation, we show the importance of our three system components and compare them with other similar approaches.

\section{Related Work}
Attention models increase the interpretability of deep neural networks internal representations by capturing where the model is focusing its attention when performing a particular task. Sharma et al. \cite{DBLP:journals/corr/SharmaKS15} proposed a Soft-Attention LSTM model to selectively focus on parts of the video frames and classify videos after taking a few glimpses. As far as we know, unlike similar task such as action recognition, there has been relatively little work that explores spatial-attention for emotion recognition. Zhang et al.  \cite{DBLP:journals/corr/abs-1711-09550} proposed attention based on fully convolutional neural network for audio emotion recognition which helped the model to focus on the emotion-relevant regions in speech spectrogram.

For capturing temporal dependencies between video frames in video classification, recurrent neural networks (RNN), particularly long short-term memory (LSTM) \cite{Hochreiter:1997:LSM:1246443.1246450} have been applied in numerous papers \cite{Chen:2015:MDE:2808196.2811638},\cite{Liu:2018:MBE:3242969.3264989},\cite{Lu:2018:MSF:3242969.3264992}. However, the accuracy on video classification with these RNN-based methods were the same or worse, which may indicate that long-term temporal interactions are not crucial for video classification. Karpathy et al. \cite{KarpathyCVPR14} explored multiple approaches based on pooling local spatio-temporal features extracted by CNNs from video frames. However, their models display only a  modest improvement compared to single-frame models. In the emotion detection task, Knyazev et al. \cite{knyazev2018leveraging} exploited several aggregation functions (e.g., mean, standard deviation) allowing the incorporation of temporal features. Inspired by the attention mechanism in \cite{lin2017structured}, our proposed method explores the potential use of a self-attentive network in emotion recognition. 

\section{Proposed Model}
\begin{figure*}[!t]
\centering
\includegraphics[width=\textwidth]{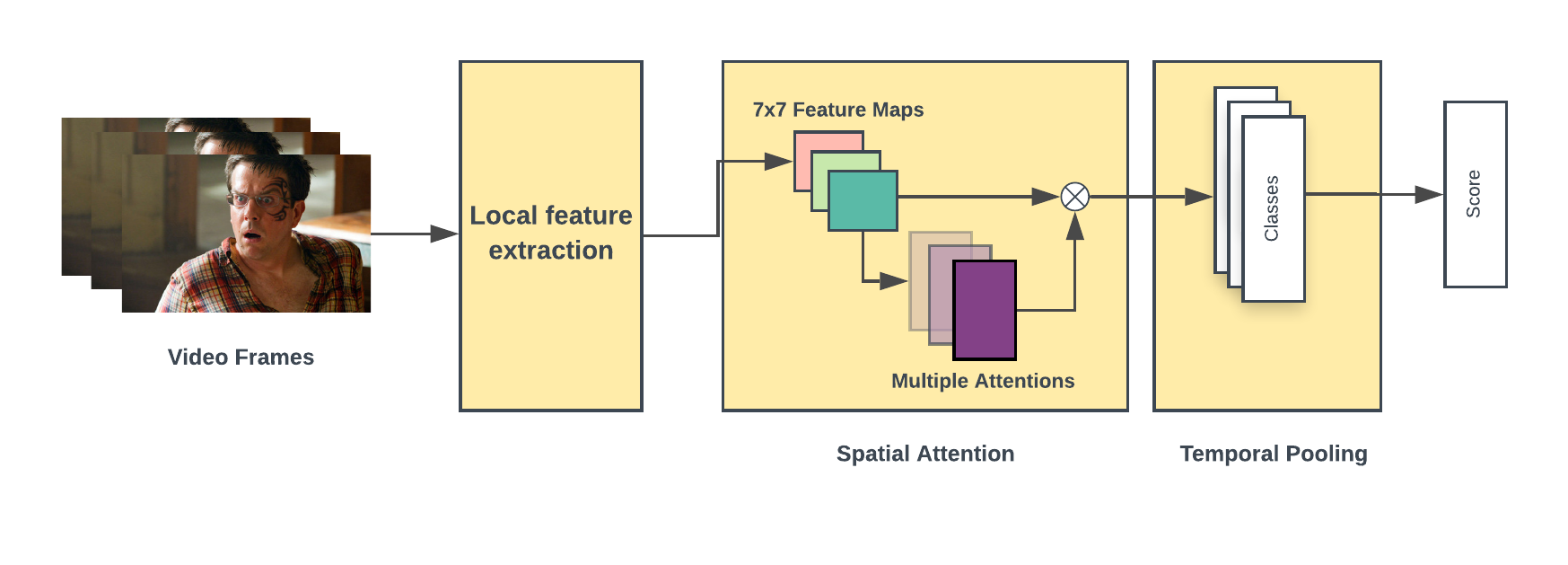}
\caption{The CNN takes the video frame as its input and produces local features. Using the local features, the multi-head attention network computes the weight importance of each local image feature. The aggregated representation is computed by multiplying multi-head attention output and the local image features. This representation is then propagates through temporal softmax pooling to extract global features over the entire video.}
\label{fig_1}
\end{figure*}

We now describe our method based on spatial attention and temporal softmax pooling for the task of emotion recognition in videos. We broadly consider three major parts: local feature extraction, local feature aggregation and global feature classification. 
The overall model architecture is shown in Fig.~ \ref{fig_1}.
The local feature extraction uses a pre-trained CNN, the spatial feature aggregation is implemented using an attention network, and the temporal feature classification uses a softmax pooling layer. 
Given a video sample $\bold S_i$ and its associated emotion $y_i \in \mathbb{R}^E$, we represent the video as a sequence of $F$ frames $ [\bold X_{0,i},\bold X_{1,i},..,\bold X_{F,i} ]$ of size $W\times H \times 3$. 

\subsection{Local Feature Extraction}
We use the VGG-16 architecture with the pre-trained VGG-Face Model\cite{Parkhi15} for extracting an independent description of a face on each frame in the video. For a detailed procedure of face extraction, see the experimental results in section~\ref{sec::experiments}.
For a given frame $\bold X$ of a video, we consider the feature map produced by the last convolutional layer of the network as representation. This feature map has spatial resolution of $L=H/16 \times W/16$ and $D$ channels. We discard the spatial resolution and reshape the feature map as a matrix ${\bold R}$ composed of $L$ D-dimensional local descriptors (row vectors). 
\begin{equation}
\bold R = VGG_{16}(\bold X)
\end{equation}

These descriptors will be associated to a corresponding weight and used for the attention mechanism.

\subsection{Spatial Attention}
For the spatial attention we rely on the self-attention mechanism \cite{vaswani2017attention}, which aggregates a set of local frame descriptors $\bold R$ into a single weighted sum $v$ that summarizes the most important regions of a given video frame: 

\begin{equation}
v = a \bold R,
\end{equation}
where $a$ is a row vector of dimension $L$, which defines the importance of each frame region.
The weights $a$ are generated by a two-layers fully connected network that associates each local feature (row of $\bold R$) to a corresponding weight:
\begin{equation}
{a} = softmax({w}_{s2}tanh(\bold W_{s1} \bold R^\top)).
\end{equation}
$\bold W_{s1}$ is then a weight matrix of learned parameters with shape $U \times D$ and ${w}_{s2}$ is a vector of parameters with size $U$. 
The softmax function ensures that the computed weights are normalized, i.e. sum up to 1.

This vector representation usually focuses on a specific region in the facial feature, like the mouth. However, it is possible that multiple regions of the face contain different type of information that can be combined to obtain a better idea of the person emotional state. Based on \cite{lin2017structured}, in order to represent the overall emotion of the facial feature, we need multiple attention units that focus on different parts of the image. 
For doing that, we transform ${w}_{s2}$ into a matrix $ \bold W_{s2}$ of size $R \times L$, in which every row represents a different attention:
\begin{equation}
{\bold A} = softmax(\bold W_{s2}tanh(\bold W_{s1}\bold R^\top)).
\end{equation}
Here the softmax is performed along the second dimension of its input.
In the case of multiple attention units, the aggregated vector $v$ becomes a matrix $D \times N$ in which each row represents a different attention. This matrix will be then flattened back to a vector $v$ by concatenating the rows in a single vector.
Thus, with this approach, a video is now represented as a $F \times (ND)$ matrix $\bold V$ in which every row is the attention based description of a video frame.
To reduce the possible overfitting of the multiple attentions, similarly to  \cite{lin2017structured} we regularize $\bold A$ by computing Frobenius norm of matrix $(\bold A \bold A^\top - I)$ and adding it to the final loss. 
This enforces diversity among the attentions and resulted very important in our experiments for good results.


\subsection{Temporal Pooling}
After extracting the local features and aggregating them using the attention mechanism for each individual frame, we have to take into account frame features over the whole video. As the length of a video can be different for each example, we need an approach that support different input lengths. The most commonly used approaches are average and max pooling; however, these techniques assume that every frame of the video has the same importance in the final decision (average pooling) or that only a single frame is considered as a general representation of the video (max pooling). In order to use the best of both techniques,  we use an aggregation based on softmax, which can be considered a generalization of the average and max pooling. In practice, instead of performing the classical softmax on the class scores, to transform them in probabilities to be used with cross-entropy loss, we compute the softmax on the class probabilities and the video frames jointly.
Given a video sample $ \bold S$, after feature extraction and spatial attention we obtain a matrix $\bold V$ in which each row represents the features of a frame. These features are converted into class scores thorough a final fully connected layer $\bold O = {\bold W_{sm} \bold V}$. In this way $\bold O$ is a $F\times E$ matrix in which an element $o_{c,f}$ is the score for class $c$ of the frame $f$.
We then transform the scores over frames and classes in probabilities with a softmax: 
\begin{equation}
p(c,f|\bold S) = \frac{exp(o_{c,f})}{\sum_{j,k} exp(o_{j,k})}.
\end{equation}
In this way, we obtain a joint probability on class $c$ and frame $f$.
From this, we can marginalize over frames $
p(c|\bold S) = \sum_f p(c,f|S)
$
and obtain a classification score that can be used in the training process using cross-entropy loss:
\begin{equation}
\mathcal{L}_{CE}= \sum_i -\log(p({y_i}|\bold S_i)).
\end{equation}
On the other hand, the same representation can be marginalized over classes $p(f|\bold S) = \sum_c p(c,f|\bold S)$. In this case, it will give us information about the most important frames of a given video (see Fig. \ref{fig_2}). This mechanism looks very similar to attention, but it has the advantage to not require an additional network to compute the attention weights. This can be important in cases for which the training data is limited and adding a sub-network with additional parameters to learn could lead to overfitting.
In this case, the weight associated to each frame and each class are computed as a softmax of the score obtained.

\section{Experiments}
\label{sec::experiments}
\subsection{Data Preparation}
We evaluate our models based on AFEW database, which is used in the audio-video sub-challenge of the EmotiW\cite{Dhall:2012:CLR:2360764.2361172}. AFEW is collected from movies and TV reality shows, which contains 773 video clips for training and 383 for validation. We extract the frame faces using the dlib\cite{king2009dlib} detector for achieving effective facial images. Then faces are aligned to a frontal position and stored in a resolution of $256 \times 256$ pixels, ready to be passed to VGG16.

\subsection{Training Details} \label{training}
To overcome overfitting during training we sampled 16 random frames form the video clips. Before feeding the facial image to the network we applied data augmentation: flipping, mirroring and random cropping of the original image. We set weight decay penalty to $0.00005$ and use SGD with momentum and warm restart \cite{DBLP:journals/corr/LoshchilovH16a} as optimization algorithm. All models are fine-tuned for 30 epochs, but we use a learning rate of $0.00001$ for the backbone CNN parameters and $0.1$ for the rest of the parameters. 


\begin{figure*}[!t]
\centering
\includegraphics[width=\textwidth]{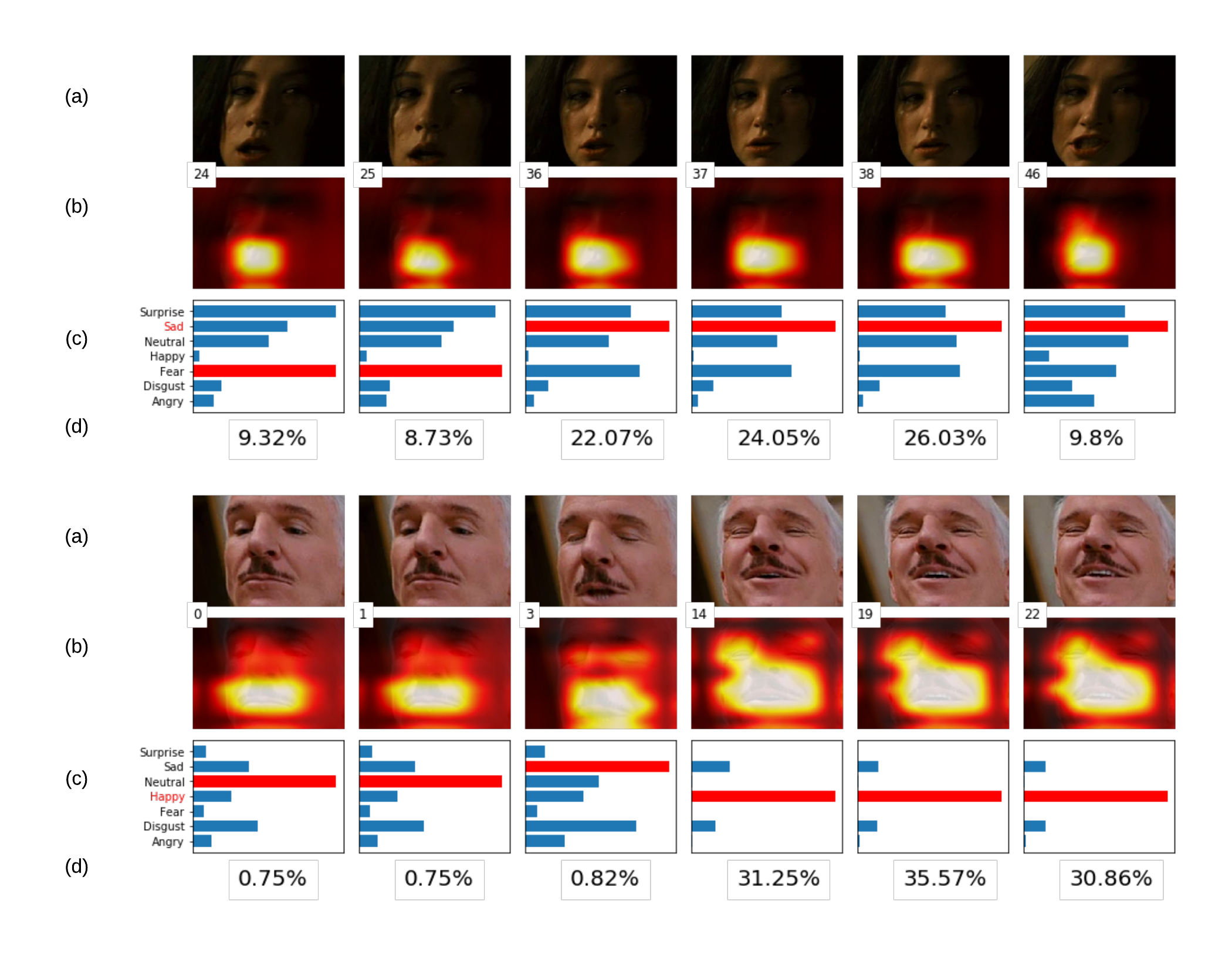}
\caption{Video frames for a few time-steps for an example of sadness and anger. (a) Original video frames (b) Image regions attended by the spatial attention mechanism. Whiter regions represent the most important parts of the face to recognize a certain emotion for the attention. (c) Emotion probability for each frame. The red bar shows the selected emotion. (d) Temporal importance for selected frames. To make those values more meaningful they have been re-normalized between 0 and 100\%.}
\label{fig_2}
\end{figure*}

\subsection{Spatial Attention}
Table \ref{table:2} reports the accuracy based on the AFEW validation dataset. We compare our softmax-based temporal pooling with different configurations of attention by varying the number of attention models and the used regularization. Using just one attention does not helps to improve the overall performance. This is probably due to the fact that a single attention usually focuses on a specific part of the face, like a mouth. However, there can be multiple regions in a face that together forms the overall emotion of the person. Thus, we evaluate our model with 2 and 4 attentions. 
The best results are obtained with 2 attention models and a strong regularization that enforces the models to focus on different parts of the face.
We observe that, adding more than two attentions do not improve the overall performance.
This is probably due overfitting. 
We also compare with our re-implementation of cluster attention with shifting operation (SHIFT) \cite{DBLP:journals/corr/abs-1711-09550}, but results are lower than our approach.
In Fig. \ref{fig_2}(b) we show that the throughout the frames, model not only captures the mouth, which in this case is the most important part for detecting the emotion but also in the first three frames focuses on the eyes as well.


\begin{table}[h!]
\centering
\caption{We evaluate the performance of the proposed spatial attention and compare it with a baseline model without it (first row). TP is our temporal softmax, while SA is the spatial attention. The second column reports the number of attention models used and the third column the amount of regularization as described in sec. \ref{training}. Finally the last column reports the accuracy on the validation set. }
\label{table:2}
 \begin{tabular}{l c c c} 
 Model & \# Att. & Reg. & ACC \\ [0.5ex] 
 \hline
 $\mathrm{VGG}_{16}+ TP$ & - & - & 46.4\% \\ 
  $\mathrm{VGG}_{16}+ TP + SHIFT$ & 2 & - & 45.0\% \\ 
 $\mathrm{VGG}_{16}+ TP + SA$ & 1 & 0 & 47.6\% \\
 $\mathrm{VGG}_{16}+ TP + SA$ & 2 & 0.1 & 48.9\% \\
 $\mathrm{VGG}_{16}+ TP + SA$ & 2 & 1 & \textbf{49.0}\% \\
 $\mathrm{VGG}_{16}+ TP + SA$ & 4 & 0.1 & 48.3\% \\
 $\mathrm{VGG}_{16}+ TP + SA$ & 4 & 1 & 48.6\% \\
\end{tabular}
\end{table}

\subsection{Temporal Pooling}
In this section we compare the performance of different kind of temporal pooling. 
The simplest approach is to consider each video sample $i$ frame independent from the others $p(c|\bold S_i)=\prod_f p(c,f|\bold X_{f,i})$ and associating the emotion class $c$ of a video to all its frames. In this case the loss becomes:
\begin{equation}
    \mathcal{L}_{CE}=\sum_i -\log(p(c|\bold S_i)) = \sum_i \sum_f -log(p(c,f|\bold X_{f,i})),
\end{equation}
which can be computed independently from each frame. In this way we can avoid to keep in memory at the same time all the frames of a video. However assuming that each frame of the same video is independent from the others, it is a very restrictive assumption and it is in contrast with the common assumption used in learning of identically independently distributed samples. We notice that this approach is equivalent to perform an average pooling (VGG+AVG) on the scoring function before the softmax normalization. This can explain the lower performance of this kind of pooling.

In Table \ref{table:1} we report results of different pooling approaches. We report results for \cite{Liu:2018:MBE:3242969.3264989} in which they use VGG16 with an LSTM model to aggregate frames (LSTM). We compare it with a VGG16 model trained with average pooling (AVG) and our softmax temporal pooling (TP). Finally we also consider the model with our temporal pooling and spatial attention (TP+SP). 
It is interesting to note that our model, even if not explicitly reasoning on the temporal scale, (.i.e every frame is still computed independently, but then the scores are normalized with the softmax) outperforms a model based on LSTM a state-of-the-art recurrent neural network. This suggest that for emotion recognition it is not really important the sequentiality of the facial postures, but the presence of certain key patterns.  

\begin{table}[h!]
\centering
\caption{We compare our softmax temporal aggregation (VGG+TP) with the approach of \cite{Liu:2018:MBE:3242969.3264989} based on recurrent neural networks (VGG+LSTM) and average pooling (VGG+AVG). Our temporal pooling is already slightly better than a more complex approach based on a recurrent network that keep memory of the past frames. Finally, if we add the spatial attention (VGG+TP+SA), we obtain a gain of almost 3 points. }
\label{table:1}
 \begin{tabular}{l c} 
 Model & ACC \\ [0.5ex] 
 \hline
 $\mathrm{VGG}_{16}+ LSTM$ \cite{Liu:2018:MBE:3242969.3264989} & 46.2\% \\
 $\mathrm{VGG}_{16}+ AVG$ & 46.0\% \\
 $\mathrm{VGG}_{16}+ TP$ & 46.4\% \\
 $\mathrm{VGG}_{16}+ TP + SA$ & \textbf{49.0}\% \\
\end{tabular}
\end{table}

\section{Conclusion}
In this paper we have presented two simple strategies to improve the performance of emotion recognition is video sequences. In contrast to previous approaches using recurrent neural networks for the temporal fusion of the data, in this paper we have shown that a simple softmax pooling over the emotion probabilities, that selects the most important frames of a video, can lead to promising results. Also, to obtain more reliable results, instead of fusing multiple sources of information or multiple learning models (e.g. CNN+C3D), we have used a multi-attention mechanism to spatially select the most important regions of an image. 
For future work we plan to use similar techniques to integrate other sources of information such as audio.
%
%
%
\bibliographystyle{splncs04}
\bibliography{refs}

\end{document}